\documentclass[runningheads]{llncs}

\usepackage{eccv}
\usepackage{multirow}
\usepackage{eccvabbrv}

\usepackage{graphicx}
\usepackage{booktabs}
\usepackage[linesnumbered,ruled,vlined]{algorithm2e}
\usepackage{amsmath}
\usepackage[accsupp]{axessibility}  % Improves PDF readability for those with disabilities.

\usepackage{hyperref}

\usepackage{orcidlink}
\usepackage{makecell}

\begin{document}

\title{HQ-DM: Single Hadamard Transformation-Based Quantization-Aware Training for Low-Bit Diffusion Models} 

\titlerunning{HQ-DM}

\author{Shizhuo Mao$^*$\orcidlink{0009-0002-1852-2176} \and
Hongtao Zou$^*$\orcidlink{0009-0009-4381-506X} \and
Qihu Xie\orcidlink{0009-0004-0519-1157} \and
Song Chen\orcidlink{0000-0003-0341-3428} \and
Yi Kang\inst{\dagger}\orcidlink{0000-0002-5487-6855}}

\authorrunning{S.~Mao et al.}

\institute{University of Science and Technology of China, Hefei, China \\
\email{\{msz123,zouht49,xieqihu\}@mail.ustc.edu.cn and \{songch,ykang\}@ustc.edu.cn}\\
}

\maketitle
\def\thefootnote{}
\footnotetext[0]{$^*$These authors contributed equally. \\ $^\dagger$Corresponding author.}

% \section*{Acknowledgements}
% Please insert your acknowledgments here.
\begin{abstract}
Diffusion models have demonstrated significant applications in the field of image generation. However, their high computational and memory costs pose challenges for deployment. Model quantization has emerged as a promising solution to reduce storage overhead and accelerate inference. Nevertheless, existing quantization methods for diffusion models struggle to mitigate outliers in activation matrices during inference, leading to substantial performance degradation under low-bit quantization scenarios. To address this, we propose \textbf{HQ-DM}, a novel Quantization-Aware Training framework that applies Single Hadamard Transformation to activation matrices. This approach effectively reduces activation outliers while preserving model performance under quantization. Compared to traditional Double Hadamard Transformation, our proposed scheme offers distinct advantages by seamlessly supporting INT convolution operations while preventing the amplification of weight outliers. For conditional generation on the ImageNet $256\times256$ dataset using the LDM-4 model, our W4A4 and W4A3 quantization schemes improve the Inception Score by 12.8\% and 467.73\%, respectively, over the existing state-of-the-art method.
\end{abstract}    
\section{Introduction}
\label{sec:intro}

In recent years, deep generative models have exhibited remarkable capabilities across multiple modalities, including image~\cite{Karras2019StyleGAN,Rombach2022LDM,podell2024sdxl}, speech~\cite{Oord2016WaveNet,shen2024naturalspeech,liu2024audioldm}, and text~\cite{Brown2020GPT3,touvron2023llama}. Among them, diffusion models~\cite{Ho2020DDPM,Rombach2022LDM} have emerged as a mainstream generative framework due to their superior generative quality and stable training dynamics. However, the generation process of diffusion models relies on an iterative, multi-step denoising sampling procedure~\cite{Ho2020DDPM}, resulting in prolonged inference latency and high memory consumption that hinder their deployment in real-time or resource-constrained scenarios.

Model quantization~\cite{Li2023QDiffusion,Ryu2025DGQ,Gao2025MoDiff} is an effective model compression method that maps high-precision numerical values to low-bit representations, thereby achieving significant memory savings and computational acceleration. Existing quantization approaches are typically categorized into two types: Post-Training Quantization (PTQ) and Quantization-Aware Training (QAT)~\cite{zhu2024survey}. PTQ~\cite{librecq,shang2023post,weiqdrop} performs offline discretization of weights and activations after training, offering ease of deployment and low implementation cost, but it may suffer from notable accuracy degradation in complex generative tasks. In contrast, QAT~\cite{he2024efficientdm,Esser2020LSQ} introduces quantization noise during training, enabling the model to adapt to low-precision arithmetic through gradient-based optimization, thereby achieving a more favorable balance between model accuracy and quantization efficiency.

\begin{figure*}[t]               
  \centering                       
  \includegraphics[width=1.0\textwidth]{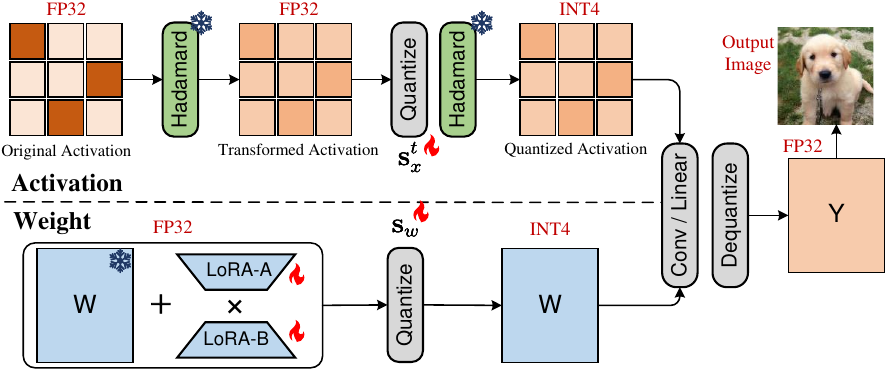}
  \caption{Overview diagram of the HQ-DM distillation framework, illustrating the application of Single Hadamard transformation before activation quantization to reduce outliers, followed by hardware-efficient implementation of matrix multiplication and convolution operations using quantized integer matrices.} 
  \label{fig:overflow}               
\end{figure*}

Nevertheless, applying quantization techniques directly to diffusion models presents significant and unique challenges. Unlike traditional classification or generative tasks, the activations of the noise prediction network in diffusion models exhibit strong non-stationarity during the multi-step denoising process: the scale of input noise, feature structures, and statistical properties vary substantially across timesteps, leading to continuously shifting activation distributions throughout sampling~\cite{Li2023QDiffusion,nichol2021improved}. Moreover, the prevalent long-tailed and outlier characteristics in diffusion models further enlarge the activation dynamic range, making them susceptible to significant clipping and rounding errors under low-bit quantization~\cite{lee2025dmq}. More critically, these errors accumulate over multiple denoising steps, resulting in cumulative and amplified degradation of generation quality.
Consequently, the non-stationary and outlier behaviors of activations constitute the key bottlenecks that hinder accurate quantization of diffusion models, calling for systematic solutions to mitigate their impact.

Several recent studies have begun to explore the challenges of temporal quantization in diffusion models. For instance, Li et al.~\cite{Li2023QDiffusion} identified the issues of temporal variation in activation distributions and the accumulation of quantization errors, and proposed a timestep-aware calibration mechanism to mitigate performance degradation in PTQ. Similarly, So et al.~\cite{So2023TDQ} further analyzed how temporal activation dynamics influence quantization interval selection. EfficientDM~\cite{he2024efficientdm} introduced a LoRA-based distillation scheme to effectively enhance quantized model performance, thereby bringing efficient quantization-aware training into diffusion model quantization. However, due to the lack of further methods to handle activation outliers, existing approaches still suffer from significant performance degradation under low-bit quantization.  

Hadamard transformation is an orthogonal transformation capable of mapping outliers into another linear space, a property that can be leveraged to address the issue of activation outliers that hinder quantization~\cite{ashkboos2024quarot,xi2023training}. Specifically, QuaRot~\cite{ashkboos2024quarot} introduces the Hadamard transformation into PTQ for LLMs, yet the adjustment of quantization factors following this transformation remains relatively under-explored. Xi et al.~\cite{xi2023training} incorporate the Hadamard transformation into quantization-aware training but do not account for the optimization of time-step-dependent quantization factors. Moreover, due to its adoption of the Double Hadamard Transformation, it lacks support for integer convolution computation modules, and the offline-generated Hadamard-transformed weight matrices amplify quantization errors.

To tackle these challenges, this paper introduces a Single Hadamard \\ Transformation-Based Quantization-Aware Training (QAT) method for diffusion models, which mitigates outlier effects via the Hadamard transformation and integrates LoRA-based weight fine-tuning with timestep-wise learnable quantization factors. This design significantly alleviates the accumulation of quantization errors under data-free fine-tuning conditions. Experimental results show that our method substantially improves generation quality at the W4A4 / W4A3 quantization level, demonstrating its practicality and strong generalization potential for diffusion model quantization. Overall, the main contributions of this paper are summarized as follows:

(1) We introduce a Single Hadamard Transformation-based distillation framework that reduces activation outliers prior to quantization through Hadamard transformation, thereby enhancing diffusion model performance under low-bit quantization settings.

(2) We propose the Single Hadamard Transformation, which is more \\ implementation-friendly than conventional approach. This transformation addresses two critical limitations of the conventional Hadamard transformation: its incompatibility with integer (INT) convolution operations and its tendency to generate additional outliers in weight matrices during computation.

(3) HQ-DM demonstrates exceptional performance across multiple datasets and models, surpassing existing SOTA methods, particularly in low-bit activation quantization scenarios. Under W4A3 quantization with the LDM-4 model on ImageNet $256\times256$, HQ-DM achieves a 467.73\% improvement on IS score, while under W4A4 quantization it delivers a 12.8\% enhancement on IS score compared to EfficientDM.

%-------------------------------------------------------------------------

\section{Related Work}
\label{sec:related work}

\textbf{Quantization of Diffusion Models.} Diffusion models represent a class of generative models that learn data distributions through a progressive denoising process, typically formulated as a Markov chain that gradually transforms random noise into structured samples. Recently, quantization has been extended to diffusion models to reduce their prohibitive computational and memory overhead. Early PTQ-based approaches such as PTQ4DM~\cite{shang2023post} and Q-Diffusion~\cite{Li2023QDiffusion} leverage data-free calibration by extracting intermediate activations from the denoising process. 
More advanced methods, including PTQD~\cite{he2023ptqd}, TFMQ-DM~\cite{huang2024tfmqdm} and APQ-DM~\cite{wang2024apqdm}, enhance quantization precision through improved calibration and temporal feature modeling. 
However, PTQ-based strategies still exhibit severe performance drops at 4-bit or lower precision due to activation outliers and the non-stationary feature dynamics across timesteps. 
To address this issue, QAT-based methods like EfficientDM~\cite{he2024efficientdm} and QuEST~\cite{wang2025quest} incorporate fine-tuning to recover generation quality under low-bit settings. Despite the advancements achieved by these works in diffusion model quantization, these quantization schemes lead to severe performance degradation under lower-bit quantization scenarios.

\textbf{Activation Outliers and Smoothing Techniques for Model Quantization.}
Post-training quantization (PTQ) is an effective approach to reduce model size and inference cost without retraining. 
However, neural activations often follow heavy-tailed distributions, where a few large-magnitude outliers dominate the tensor range and severely degrade quantization accuracy~\cite{wei2022outlier}. 
Early works mitigate this by clipping or architectural transformations such as Outlier Channel Splitting (OCS)~\cite{zhao2019improving}, while recent methods for large models introduce algebraically equivalent scaling to smooth activation distributions without retraining. 
Notably, SmoothQuant~\cite{xiao2023smoothquant} migrates quantization difficulty from activations to weights via scaling, proving that explicit outlier handling is the key to achieving low-bit quantization.
Diffusion models exhibit more complex outlier behaviors: activation statistics vary significantly across timesteps, and quantization errors can accumulate during iterative denoising. To address this, diffusion-specific PTQ methods introduce timestep- or layer-aware smoothing mechanisms. 
DMQ~\cite{lee2025dmq} analyzes outlier sources in diffusion networks and applies diffusion-aware transformations to suppress extreme activations, while MPQ-DM~\cite{feng2025mpq} leverages mixed-precision allocation guided by outlier statistics to maintain quality under ultra-low-bit settings. 
Overall, these studies demonstrate that timestep-adaptive outlier smoothing—through equivalent scaling or precision allocation—is essential for stable and accurate quantization of diffusion models.

QuaRot~\cite{ashkboos2024quarot} introduces Hadamard transformation into Post-Training Quantization (PTQ) for large language models to mitigate outliers and enhance model performance under low-bit quantization. Similarly, Xi et al.~\cite{xi2023training} incorporate the Hadamard transformation into the training framework of large models to reduce outliers and improve performance. However, these works do not optimize time-step-dependent quantization factors as training objectives. Moreover, the proposed Double Hadamard Transformation suffers from two critical limitations: incompatibility with integer-based convolution operations and the introduction of additional outliers in weight matrices, thereby hindering its effective application in diffusion models.

\section{Background}
\paragraph{\textbf{Diffusion Model}}
Diffusion Models (DMs) learn to reverse a gradual noising process, transforming a simple Gaussian prior into complex data distributions through iterative denoising~\cite{Ho2020DDPM,Rombach2022LDM}.
\begin{figure*}[t]               
  \centering                       
  \includegraphics[width=1.0\textwidth]{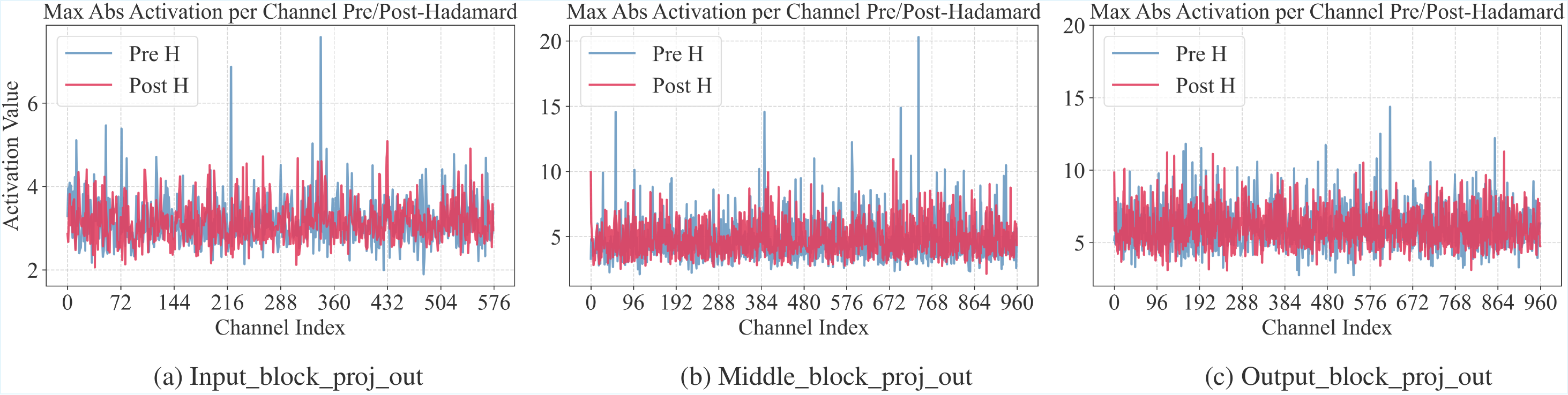}
  \caption{Channel-wise analysis of outlier distribution in activations from the input\_block, middle\_block and output\_block layer at timestep 0, comparing before and after Hadamard Transformation states. Given that the minimum abs value of each channel in the matrix is zero, the maximum value of each channel can be considered as the outlier for that channel.} 
  \label{fig:outlier}               
\end{figure*}
Formally, the forward diffusion process progressively adds Gaussian noise to a clean data sample $x_0$ over $T$ timesteps:
\begin{equation}
q(x_t \mid x_{t-1}) = \mathcal{N}(x_t; \sqrt{1 - \beta_t}\, x_{t-1}, \beta_t I)
\end{equation}

where $\beta_t$ controls the noise variance at each timestep. 
During training, a denoising network $\epsilon_\theta(x_t, t)$ is optimized to predict the added noise by minimizing the simplified objective:
\begin{equation}
\mathcal{L}_{\text{simple}} = 
\mathbb{E}_{x_0, t, \epsilon} 
\left[
\| \epsilon - \epsilon_\theta(\sqrt{\bar{\alpha}_t} x_0 + \sqrt{1 - \bar{\alpha}_t}\, \epsilon, t) \|^2
\right]
\end{equation}
where $\bar{\alpha}_t = \prod_{i=1}^{t} (1 - \beta_i)$.
To accelerate sampling, several variants have been proposed. 
DDIM~\cite{Song2021DDIM} introduces deterministic sampling to reduce inference steps, while 
Latent Diffusion Models (LDMs)~\cite{Rombach2022LDM} perform diffusion in a compressed latent space, greatly improving efficiency. 

\paragraph{\textbf{Model Quantization}}
Quantization refers to the process of mapping high-precision numbers to low-bit representations. A typical quantization method is linear mapping quantization, which maintains a stable quantization step size and uniformly maps values within the dynamic range to integer numbers. The formula for symmetric quantization from floating-point numbers to N-bit integers is as follows:
\begin{align}
\mathbf{X_{\text{int}}} = \text{clamp}\left( \left\lfloor \frac{\mathbf{X_{\text{fp}}}}{\mathbf{S}} \right\rceil, min, max \right), \quad \mathbf{\hat{X}} = \mathbf{S} \cdot \mathbf{X_{\text{int}}}
\end{align}
where $\lfloor \cdot \rceil$ represents the rounding function, and $\text{clamp}$ denotes truncation within the integer range representable by N-bit precision. $ \mathbf{S} $ represents the quantization scale factor, which is a trainable parameter in this paper. Asymmetric quantization involves adding an offset (zero-point) after dividing $ \mathbf{X} $ by the scale factor $ \mathbf{S} $, which alters the dynamic range of the value distribution. To overcome the non-differentiability of the rounding and clamping functions during the QAT process, we adopt the Straight-Through Estimator (STE) convention:
\begin{equation}
    \frac{\partial \text{Round}(x)}{\partial x} = 1, \quad
    \frac{\partial \text{Clamp}(x)}{\partial x} = \mathbb{I}_{[\text{min}, \text{max}]}(x)
\label{eq:clamp_grad}
\end{equation}
Outliers hold significant implications in model quantization, referring to large values that substantially exceed the magnitude of other values. Due to the presence of these outliers, it becomes challenging to simultaneously preserve information from both normal values and outliers after quantization, consequently leading to substantial performance degradation in low-bit quantization scenarios. Compared to weight values, activations typically exhibit more frequent occurrences of outliers~\cite{dettmers2022gpt3, yao2022zeroquant, wei2022outlier}. Therefore, to maintain model performance under low-bit quantization, it is crucial to mitigate the impact caused by these outlier values.

\paragraph{\textbf{Hadamard Transformation}}
An orthogonal matrix is defined as a real square matrix \(\mathbf{Q}\) satisfying \(\mathbf{Q}\mathbf{Q}^{T}=\mathbf{I}\). The linear transformation corresponding to an orthogonal matrix preserves both vector lengths and angles between vectors, making it particularly suitable for performing linear transformations on matrices. A Hadamard matrix represents a special class of orthogonal matrices, satisfying the following recurrence relation:
\begin{equation}
\mathbf{H}_0 = [1], \quad \mathbf{H}_k = \frac{1}{\sqrt{2}} \begin{bmatrix} \mathbf{H}_{k-1} & \mathbf{H}_{k-1} \\ \mathbf{H}_{k-1} & -\mathbf{H}_{k-1} \end{bmatrix}
\end{equation}
The matrix $\mathbf{H}_k$ is of size $2^k \times 2^k$ and possesses both orthogonality and rotational symmetry. Consequently, it satisfies  
\begin{align}
\mathbf{H}_k \mathbf{H}_k^\top = \mathbf{I} \quad \text{and} \quad \mathbf{H}_k = \mathbf{H}_k^\top
\end{align}
It can be observed that the orthonormal Hadamard matrix is defined as
\begin{align}
\mathbf{H}_k = 2^{-k/2} \cdot \mathbf{H}_k^{\text{raw}}
\end{align}
where \(\mathbf{H}_k^{\text{raw}}\) is the unnormalized Hadamard matrix with all entries in \(\{+1, -1\}\). This decomposition is crucial for integer-only quantized inference, ensuring that the matrix multiplication after transformation can be performed entirely within integer arithmetic. For any matrix-vector multiplication of the form $\mathbf{H}_k \mathbf{x}$, the computational complexity is $\mathcal{O}(k \cdot 2^k)$ operations. 
The Hadamard transformation is a linear transformation that uses a Hadamard matrix as its transform kernel. 

The order $k$ of the base Hadamard matrix $\mathbf{H}_k$ is directly related to its outlier-diffusion capability: larger $k$ leads to stronger diffusion of dominant outliers across all coordinates. Fundamentally, the Hadamard transformation maps input vectors into a linear subspace with fewer extreme values, thereby significantly improving model performance under low-bit quantization.

\section{Method}

\begin{figure*}[t]               
  \centering                       
  \includegraphics[width=0.85\textwidth]{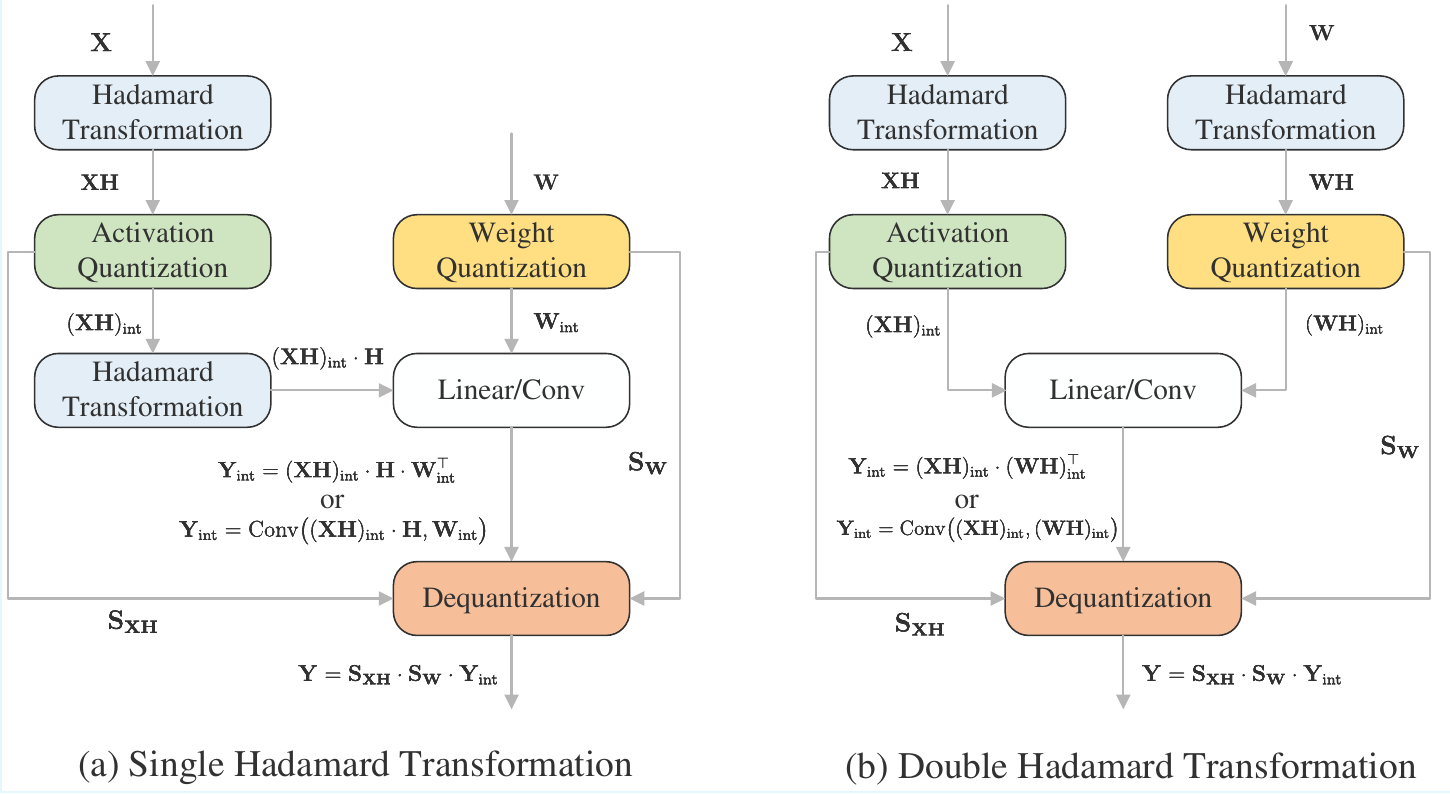}
  \caption{Schematic diagrams of Single Hadamard Transformation (a) and Double Hadamard Transformation (b). The key distinction lies in whether the offline Hadamard transformation is applied to the weights. } 
  \label{fig:singledouble}               
\end{figure*}

In the context of low-bit quantization for diffusion models, we employ the Hadamard transformation to suppress outliers and thereby enhance model performance under aggressive quantization. In this section, we present our Hadamard-based LoRA fine-tuning framework—termed \textbf{HQ-DM} (Single Hadamard Transformation-Based Quantization Aware Training on Diffusion Models)—which applies quantization to both activations and weights, as illustrated in Figure~\ref{fig:overflow}.
\subsection{LoRA-based Distillation}
Inspired by EfficientDM~\cite{he2024efficientdm}, we retain the LoRA-based distillation framework. The core idea of Low-Rank Adaptation (LoRA) is to freeze the original pre-trained weight and only train a pair of low-rank update matrices that are adapted to it~\cite{hu2022lora,dettmers2023qlora}. This strategy drastically reduces the number of trainable parameters while preserving a sufficiently expressive parameter space, as the adaptation is concentrated on the dominant principal components of the weight updates.
Formally, consider a linear transformation $\mathbf{Y} = \mathbf{X} \cdot \mathbf{W}$, where $\mathbf{X} \in \mathbb{R}^{T \times C_i}$ and $\mathbf{W} \in \mathbb{R}^{C_i \times C_o}$. Under LoRA, the adapted weight matrix becomes
\begin{align}
\mathbf{W}' = \mathbf{W} + \mathbf{B} \cdot \mathbf{A}
\end{align}
where $\mathbf{B} \in \mathbb{R}^{C_i \times r}$ and $\mathbf{A} \in \mathbb{R}^{r \times C_o}$ are trainable low-rank matrices with $r \ll \min(C_i, C_o)$.
In the distillation process, at denoising timestep $t$, the same noisy input $\mathbf{x}_t$ is fed into both the full-precision (FP) teacher model and the quantized student model, and the mean squared error (MSE) between their outputs is minimized:
\begin{align}
\mathcal{L}_{\text{distill}} = \mathbb{E}_{\mathbf{x}_t, t} \left[ \left\| \mathbf{\epsilon}_{\text{FP}}(\mathbf{x}_t, t) - \mathbf{\epsilon}_{\text{quant}}(\mathbf{x}_t, t) \right\|_2^2 \right]
\end{align}
Prior work has shown that in diffusion models, the activation distributions vary significantly across timesteps. To address this, LSQ~\cite{Esser2020LSQ,he2024efficientdm} introduces timestep-aware quantization scales and optimizes them independently for each denoising step. We adopt this strategy unchanged for both weight and activation quantization in our framework.

\begin{figure*}[t]               
  \centering                       
  \includegraphics[width=1.0\textwidth]{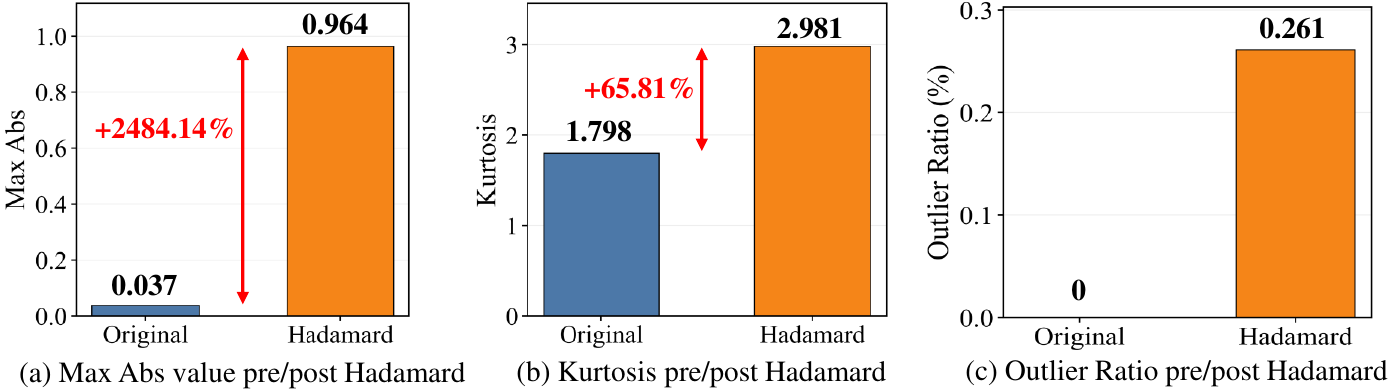}
  \caption{Comparison of max absolute value, kurtosis, and outlier ratio (values exceeding 3$\sigma$) before and after Hadamard transformation on output\_blocks0.attn2.k.weight in LDM-4. “Original” corresponds to the Single Hadamard Transformation (no weight transformation), while “Hadamard” denotes the Double Hadamard Transformation.} 
  \label{fig:weight_outlier}               
\end{figure*}
\subsection{Hadamard Transformation to smooth outliers}
We utilize Single Hadamard Transformation to reduce outliers in the activations, while avoiding the introduction of additional outliers in the weights that typically arises from the conventional Double Hadamard Transformation. As shown in Figure~\ref{fig:outlier}, the outliers in the activation values are significantly reduced after the transformation.
\paragraph{\textbf{Matrix Multiplication}}
The Hadamard transformation is applied to a 2D input matrix by multiplying the matrix to be quantized with a Hadamard matrix. Since the input channel dimension $C_i$ is not necessarily a power of two, we construct a block-diagonal Hadamard matrix $\mathbf{H} \in \mathbb{R}^{C_i \times C_i}$ composed of $m$ identical submatrices $\mathbf{H}_k$, i.e.,
\begin{equation}
\mathbf{H} = \mathrm{BlockDiag}(\underbrace{\mathbf{H}_k, \mathbf{H}_k, \dots, \mathbf{H}_k}_{m \text{ blocks}}), \quad \text{where} \quad C_i = m \cdot 2^k.
\end{equation}
This block-diagonal structure preserves orthogonality, satisfying $\mathbf{H} \mathbf{H}^\top = \mathbf{H}^2 = \mathbf{I}$. In cases where the number of input channels is smaller than the base order $k$ (e.g., during the initial projection of image features), we set $\mathbf{H} = \mathbf{I}$, effectively bypassing the transformation.
Given the orthogonality of the Hadamard matrix (\(\mathbf{H}^\top = \mathbf{H}^{-1} = \mathbf{H}\)), the original activation matrix \(\mathbf{X}\) can be exactly recovered from its transformed and quantized form via inverse transformation followed by dequantization. Specifically, let \((\mathbf{X}\mathbf{H})_{\text{int}}\) denote the integer-valued quantized representation of the transformed activations \(\mathbf{X}\mathbf{H}\), and let \(\mathbf{S}_{\mathbf{X}\mathbf{H}}\) be the corresponding floating-point quantization scale. Then the dequantized output is:
\begin{align}
\hat{\mathbf{X}} = \mathbf{S}_{\mathbf{X}\mathbf{H}} \cdot (\mathbf{X}\mathbf{H})_{\text{int}} \cdot \mathbf{H}
\end{align}
Consequently, for a linear layer \(\mathbf{Y} = \mathbf{X} \mathbf{W}\), we insert the identity \(\mathbf{H} \mathbf{H}^\top = \mathbf{I}\) to obtain:
\begin{equation}
\mathbf{Y} = \mathbf{X} \mathbf{H} \cdot \mathbf{H}^\top \mathbf{W} = (\mathbf{X} \mathbf{H}) \cdot \mathbf{H} \cdot \mathbf{W} \approx \mathbf{S}_{\mathbf{X}\mathbf{H}} \, \mathbf{S}_{\mathbf{W}} \cdot \big( (\mathbf{X}\mathbf{H})_{\text{int}} \cdot \mathbf{H} \cdot \mathbf{W}_{\text{int}} \big)
\end{equation}
In practice, the normalization factor \(2^{-k/2}\) (arising from the orthonormal definition \({\mathbf{H}}_k = 2^{-k/2} \mathbf{H}_k^{\text{raw}}\)) is absorbed into the floating-point quantization scales. As a result, the core computation \((\mathbf{X}\mathbf{H})_{\text{int}} \cdot \mathbf{H} \cdot \mathbf{W}_{\text{int}}\) involves only integer matrix multiplication, enabling efficient inference on hardware accelerators.

\paragraph{\textbf{Convolution}}
For a convolutional layer, let the input activation tensor be $\mathbf{X}_c \in \mathbb{R}^{B \times C_{\text{in}} \times h \times w}$ and the convolutional kernel be $\mathbf{W}_c \in \mathbb{R}^{C_{\text{out}} \times C_{\text{in}} \times L \times L}$, where $C_{\text{in}}$ and $C_{\text{out}}$ denote the input and output channel dimensions, and $h \times w$ and $L \times L$ are the spatial resolutions of the activations and the kernel. To apply the Hadamard transformation on convolution, we first reshape $\mathbf{X}_c$ into a 2D matrix $\widetilde{\mathbf{X}}_c \in \mathbb{R}^{(B \cdot C_{\text{in}} \cdot h) \times w}$ by merging the batch, input-channel, and height dimensions. We then compute the transformed activations as the matrix product $(\widetilde{\mathbf{X}}_c \mathbf{H}_c)$, where $\mathbf{H}_c \in \mathbb{R}^{w \times w}$ is a block-diagonal Hadamard matrix. This design ensures that the Hadamard transformation operates only along the last (width) dimension and that $\mathbf{H}_c$ is square—allowing compatibility with standardized Hadamard orders.
Due to the orthogonality of $\mathbf{H}_c$ ($\mathbf{H}_c^\top = \mathbf{H}_c^{-1} = \mathbf{H}_c$), the original activations can be recovered from the quantized transformed representation via inverse transformation and dequantization. Specifically, let $(\widetilde{\mathbf{X}}_c \mathbf{H}_c)_{\text{int}}$ denote the integer-valued quantized form of the product $\widetilde{\mathbf{X}}_c \mathbf{H}_c$, and let $\mathbf{S}_{\widetilde{\mathbf{X}}_c \mathbf{H}_c}$ be its associated floating-point quantization scale. The dequantized result is:
\begin{align}
\hat{\widetilde{\mathbf{X}}}_c = \mathbf{S}_{\widetilde{\mathbf{X}}_c \mathbf{H}_c} \cdot (\widetilde{\mathbf{X}}_c \mathbf{H}_c)_{\text{int}} \ \cdot \mathbf{H}_c
\end{align}
Consequently, the convolution operation can be rewritten using the identity $\mathbf{H}_c \mathbf{H}_c^\top = \mathbf{I}$ as:
\begin{equation}
\small
\mathbf{Y} = \mathrm{Conv}\big( (\widetilde{\mathbf{X}}_c \mathbf{H}_c) \cdot \mathbf{H}_c,\, \mathbf{W}_c \big) \approx \mathbf{S}_{\widetilde{\mathbf{X}}_c \mathbf{H}_c} \, \mathbf{S}_{\mathbf{W}_c} \cdot \mathrm{Conv}\Big( (\widetilde{\mathbf{X}}_c \mathbf{H}_c)_{\text{int}} \cdot \mathbf{H}_c,\, (\mathbf{W}_c)_{\text{int}} \Big)
\end{equation}
Similarly, in practical implementation, the normalization factor of the Hadamard matrix \(2^{-k/2}\) can be merged with the preceding floating-point quantization factors, so that the convolution operation is performed only on integer matrices.
\paragraph{\textbf{Single Hadamard Transformation}}
For any standard basis vector $\mathbf{e}_i \in \mathbb{R}^{2^k}$ (i.e., a vector with a single entry equal to 1 and all others 0), the product $\mathbf{e}_i^\top \mathbf{H}_k$ yields a constant vector:  
\begin{equation}
\mathbf{e}_i^\top \mathbf{H}_k = 2^{-k/2} \cdot \mathbf{1}_{2^k}, \quad \text{where} \quad \mathbf{1}_{2^k} = (1, 1, \dots, 1)^\top \in \mathbb{R}^{2^k}.
\end{equation} This property implies that when the input vector contains a dominant outlier---i.e., one coordinate significantly larger than the rest---the Hadamard transformation effectively maps it to a uniform (flat) vector. Such a structure is highly favorable for quantization, as the resulting equal-valued entries simplify the quantization process and improve its efficiency. However, similarly, applying a Hadamard transformation to a vector with uniformly distributed elements will increase the outliers in its distribution. For example, a vector with all elements equal to 1 (i.e. $\mathbf{1}_{2^k}$) yields $(2^{k/2}, 0, 0, \dots, 0)$ after transformation, which significantly amplifies the outliers in the resulting vector.

\begin{align}
\mathbf{1}_{2^k} \mathbf{H}_k &= 2^{k/2} \, \mathbf{e}_1^\top
\end{align}
While previous research has employed Hadamard transformation to enhance model performance, the Hadamard transformation utilized in HQ-DM differs significantly from conventional approaches: we designate our method as the Single Hadamard Transformation, in contrast to the conventional Double Hadamard Transformation (see Figure~\ref{fig:singledouble} (a) (b)).
The primary distinction lies in the computational methodology. Conventional Double Hadamard Transformation attempts to insert Hadamard transformation between the operations of $\mathbf{X}$ and $\mathbf{W}$, enabling online quantization of $\mathbf{XH}$ while performing $\mathbf{H^{\top}W}$ quantization offline. This approach introduces two critical limitations: 1) inadequate support for convolution operations in diffusion models (due to the non-applicability of matrix multiplication distributive property to convolutions, especially for multi strides), and 2) increased outliers in the weight matrix $\mathbf{W}$ when computing $\mathbf{H^{\top}W}$ with originally low-outlier $\mathbf{W}$ matrices, consequently amplifying quantization errors and degrading model performance (see Figure~\ref{fig:weight_outlier}). Our proposed Single Hadamard Transformation not only minimizes outlier introduction in weight matrices but also provides excellent convolution support. More importantly, by extracting the floating-point factor $2^{-k/2}$, the $\mathbf{H}_k^{\text{raw}}$ matrix becomes a sparse integer matrix, enabling efficient hardware implementation of intermediate results through integer matrix multiplication.

\paragraph{\textbf{Implementation-Friendly Scheme}}
Single Hadamard Transformation possesses the following characteristics: its transformation of activation matrices renders activations more amenable to quantization, thereby achieving high efficiency for low-bit integer arithmetic. In contrast to several existing approaches that employ asymmetric quantization for activations (like EfficientDM), HQ-DM adopts symmetric quantization for activations, ensuring compatibility with hardware acceleration for integer matrix multiplication. While this approach effectively reduces the available quantization bits by one compared to asymmetric quantization in extreme cases, experimental results still demonstrate the substantial advantages of HQ-DM. Furthermore, as previously discussed, the Single Hadamard Transformation enables direct execution of integer matrix multiplication and convolution operations without requiring subsequent dequantization, establishing it as a hardware-friendly scheme.

\begin{table}[!htb]
    \centering
    \begin{tabular}{lccccc}
    \hline
        Method & Bit (W/A) & IS $\uparrow$ & FID $\downarrow$ & sFID $\downarrow$ & Precision(\%)$\uparrow$ \\
    \hline
        FP & 32/32 & 365.35 & 11.22 & 7.78 & 93.68 \\
    \hline
        Q-Diffusion & 8/8 & 350.93 & 10.60 & 9.29 & 92.46 \\
        PTQD & 8/8 & 359.78 & \textbf{10.05} & 9.01 & 93.00 \\
        EfficientDM & 8/8 & 362.34 & 11.38 & 8.04 & \textbf{93.77} \\
        \textbf{HQ-DM} & 8/8 & \textbf{363.22} & 11.01 & \textbf{7.76} & \textbf{93.46} \\
    \hline
        Q-Diffusion & 4/8 & 336.80 & 9.29 & 9.29 & 91.06 \\
        PTQD & 4/8 & 344.72 & \textbf{8.74} & 7.98 & 91.69 \\
        EfficientDM & 4/8 & 351.97 & 9.96 & 7.80 & 92.48 \\
        \textbf{HQ-DM} & 4/8 & \textbf{355.59} & 10.07 & \textbf{7.14} & \textbf{93.07} \\
    \hline
        % Q-Diffusion & 8/4 & -- & -- & -- & -- \\
        PTQD & 8/4 & 1.93 & 262.01 & 111.10 & 0.54 \\
        EfficientDM & 8/4 & 252.00 & \textbf{6.75} & \textbf{8.48} & 84.90 \\
        \textbf{HQ-DM} & 8/4 & \textbf{292.76} & 7.68 & 9.93 & \textbf{88.68} \\
    \hline
        PTQD & 4/4 & 2.57 & 258.11 & 124.33 & 0.69 \\
        QuEST & 4/4 & 202.45 & \textbf{5.98} & -- & -- \\
        EfficientDM & 4/4 & 220.20 & 6.63 & 9.80 & 80.97 \\
        \textbf{HQ-DM} & 4/4 & \textbf{248.30} & 6.58 & \textbf{9.60} & \textbf{84.44} \\
    \hline
        PTQD & 4/3 & 1.62 & 297.88 & 150.46 & 0.31 \\
        EfficientDM & 4/3 & 8.74 & 99.78 & 65.79 & 27.48 \\
        \textbf{HQ-DM} & 4/3 & \textbf{49.62} & \textbf{33.47} & \textbf{12.09} & \textbf{54.87} \\
    \hline
    \end{tabular}
    \caption{Performance comparisons of fully-quantized LDM-4 models on ImageNet $256\times256$ dataset. The sampling process employs 20 denoising steps. We evaluate IS (higher is better), FID (lower is better), sFID (lower is better), Precision (higher is better) on these quantized models.}
    \label{tab:quant_imagenet}
\end{table}
\section{Experiments}
\subsection{Experiments Settings}
We conduct experiments on conditional generation using ImageNet $256\times256$~\cite{deng2009imagenet} and unconditional generation on both LSUN-Bedrooms $256\times256$ and LSUN-Churches $256\times256$~\cite{yu2015lsun}. For our backbone architectures, we employ both U-Net-based Latent Diffusion Models (LDM)~\cite{Rombach2022LDM} and Transformer-based Diffusion Models (DiT)~\cite{peebles2023scalable}. To adhere to the standard configurations of these models, the denoising process is executed using the DDIM sampler~\cite{Song2021DDIM} for LDM and the default Improved DDPM sampler~\cite{nichol2021improved} for DiT. Trainable parameters are optimized using the AdamW optimizer~\cite{kingma2015adam,loshchilovdecoupled}, with distinct initial learning rates assigned to activation quantization factors, weight quantization factors, and LoRA low-rank matrices. For evaluation, we utilize generated sample batches of 50,000 images from each model and compute metrics against reference batches. The primary evaluation metrics include Inception Score (IS)~\cite{salimans2016improved}, Fréchet Inception Distance (FID)~\cite{heusel2017gans}, sFID~\cite{nash2021generating}, Precision and Recall. We compare our results with several prominent works in the field, including Q-Diffusion~\cite{Li2023QDiffusion}, PTQD~\cite{he2023ptqd}, QuEST~\cite{wang2025quest}, TFMQ-DM~\cite{huang2024tfmqdm}, and EfficientDM~\cite{he2024efficientdm}. Our experiments cover multiple quantization bit configurations, where WxAy denotes x-bit weight quantization and y-bit activation quantization respectively. More comparisons and details can be found in Appendix.

\subsection{Experiment Results}
\paragraph{\textbf{Conditional generation on LDM}}

We evaluate the performance of HQ-DM on ImageNet $256\times256$ using LDM-4. The results in Table~\ref{tab:quant_imagenet} demonstrate that HQ-DM outperforms existing methods across these bit configurations, with this advantage becoming increasingly pronounced at lower activation quantization bit-widths. For the W4A4 configuration, HQ-DM surpasses the current SOTA method EfficientDM by 12.76\% in Inception Score. Compared to EfficientDM, HQ-DM achieves performance improvements exceeding 100\% across all four metrics under the W4A3 configuration, with approximately 99.7\% enhancement in precision, thereby highlighting HQ-DM's potential for ultra-low-bit quantization.
\paragraph{\textbf{Unconditional generation on LDM}}
\begin{table}[!htb]
    \centering
    \setlength{\tabcolsep}{4pt}
    \begin{tabular}{lccccc}
    \hline
        Method & Bit (W/A) & FID $\downarrow$ & sFID $\downarrow$ & Precision(\%) $\uparrow$   & Recall(\%) $\uparrow$  \\
    \hline
        FP & 32/32 & 7.56 & 14.06 & 51.95 & 41.18 \\
    \hline
        TFMQ-DM & 8/4 & 226.08 & 105.91 & 0.04 & 0.00 \\
        EfficientDM & 8/4 & 17.91 & 19.44 & 40.40 & \textbf{37.86} \\
        \textbf{HQ-DM} & 8/4 & \textbf{12.35} & \textbf{16.27} & \textbf{47.31} & 37.44 \\
    \hline
        TFMQ-DM & 4/4 & 234.29 & 108.42 & 0.04 & 0.00 \\
        EfficientDM & 4/4 & 16.81 & 20.81 & 41.42 & 29.66 \\
        \textbf{HQ-DM} & 4/4 & \textbf{13.15} & \textbf{17.53} & \textbf{45.32} & \textbf{36.34} \\
    \hline
        TFMQ-DM & 3/4 & 227.64 & 107.44 & 0.06 & 7.32 \\
        QuEST & 3/4 & \textbf{19.08} & 32.75 & -- & -- \\
        EfficientDM & 3/4 & 20.92 & 22.78 & \textbf{38.57} & 29.38 \\
        \textbf{HQ-DM} & 3/4 & \textbf{19.22} & \textbf{19.29} & \textbf{38.32} & \textbf{31.92} \\
    \hline
        TFMQ-DM & 2/4 & 245.85 & 130.20 & 0.06 & 0.00 \\
        QuEST & 2/4 & \textbf{24.92} & 36.33 & -- & -- \\
        EfficientDM & 2/4 & 33.68 & \textbf{27.19} & 28.76 & \textbf{20.34} \\
        \textbf{HQ-DM} & 2/4 & 28.50 & 30.17 & \textbf{32.64} & 18.76 \\
    \hline
        EfficientDM & 4/3 & 26.80 & 24.01 & 33.43 & 29.02 \\
        \textbf{HQ-DM} & 4/3 & \textbf{21.88} & \textbf{22.15} & \textbf{34.52} & \textbf{29.10} \\
    \hline
    \end{tabular}
    \caption{Performance comparisons of fully-quantized LDM-4 models on LSUN-Bedrooms $256\times256$ dataset. The sampling process employs 100 denoising steps.}
    \label{tab:quant_lsun_bedroom}
\end{table}
\begin{table}[!htb]
    \centering
    \setlength{\tabcolsep}{5pt}
    \begin{tabular}{lccccc}
    \hline
        Method & Bit (W/A) & FID $\downarrow$ & sFID $\downarrow$ & Precision(\%) $\uparrow$  & Recall(\%) $\uparrow$   \\
    \hline
        FP & 32/32 & 6.17 & 20.98 & 59.96 & 48.54 \\
    \hline
        EfficientDM & 4/4 & 12.83 & 30.04 & 52.61 & 33.69 \\
        \textbf{HQ-DM} & 4/4 & \textbf{9.64} & \textbf{28.46} & \textbf{57.10} & \textbf{35.91} \\
    \hline
        EfficientDM & 4/3 & 20.97 & 36.56 & 46.25 & 26.06 \\
        \textbf{HQ-DM} & 4/3 & \textbf{14.67} & \textbf{31.94} & \textbf{53.22} & \textbf{27.08} \\
    \hline
    \end{tabular}
    \caption{Performance comparisons of fully-quantized LDM-8 models on LSUN-Churches $256\times256$ dataset. The sampling process employs 100 denoising steps.}
    \label{tab:quant_lsun_churches}
\end{table}
Following conventional practice, we employ the LDM-4 model for unconditional image generation on LSUN-Bedrooms $256\times256$ in Table~\ref{tab:quant_lsun_bedroom}, and the LDM-8 model for unconditional generation on LSUN-Churches $256\times256$ in Table~\ref{tab:quant_lsun_churches}. On LSUN-Bedrooms, HQ-DM surpasses existing quantization schemes across multiple bit configurations, achieving FID reductions of 5.56, 3.66 and 4.92 for W8A4, W4A4 and W4A3 settings, respectively. It is noteworthy that since HQ-DM exclusively processes activations, the accuracy of the quantized model depends on weight stability. In other words, the closer the quantized weights remain to the original floating-point model, the lower the activation quantization loss achieved by HQ-DM. On LSUN-Churches, HQ-DM also outperforms EfficientDM at low quantization bit-widths. Compared to EfficientDM, HQ-DM reduces the FID by 3.19 under the W4A4 configuration and by 6.3 under W4A3. It is noteworthy that these improvements were achieved while employing hardware-friendly symmetric quantization, which typically incurs an adverse impact on performance. Across both LSUN datasets, we consistently observe that the performance gains from HQ-DM become increasingly pronounced as activation quantization bit-width decreases.
\paragraph{\textbf{Conditional generation on DiT}}
We implement the conditional generation of HQ-DM on the DiT-XL model and compare it against the EfficientDM method in Table~\ref{tab:dit_full_final}. Under the W4A4 quantization configuration, HQ-DM consistently outperforms EfficientDM across all metrics. For instance, on DiT-XL (ImageNet $256\times256$), HQ-DM achieves a relative improvement of 60.92\% in IS and 13.79\% in precision. Meanwhile, in terms of FID and sFID, HQ-DM reduces the scores by 0.79 and 3.72, respectively, compared to EfficientDM. These results demonstrate the robust scalability and effectiveness of our method on mainstream Transformer-based architectures beyond U-Net.
\begin{table}[!htb]
    \centering
    \setlength{\tabcolsep}{3pt}
    \begin{tabular}{llcccccc}
    \hline 
        Model (Dataset) & Method & Bit (W/A) & IS $\uparrow$ & FID $\downarrow$ & sFID $\downarrow$ & Precision(\%) $\uparrow$ \\
    \hline 
        \multirow{3}{*}{\makecell[c]{\textbf{DiT-XL}\\ImageNet 256}} 
        & FP & 32/32 & 477.51 & 16.86 & 11.08 & 93.25 \\
        & EfficientDM & 4/4 & 221.07 & 10.41 & 13.90 & 77.45 \\
        & \textbf{HQ-DM} & 4/4 & \textbf{355.74} & \textbf{9.62} & \textbf{10.18} & \textbf{88.13} \\
    \hline 
        \multirow{3}{*}{\makecell[c]{\textbf{DiT-XL}\\ImageNet 512}} 
        & FP & 32/32 & 451.68 & 17.71 & 8.40 & 83.60 \\
        & EfficientDM & 4/4 & 193.00 & 9.91 & 14.72 & 82.03 \\
        & \textbf{HQ-DM} & 4/4 & \textbf{278.90} & \textbf{9.04} & \textbf{11.84} & \textbf{84.34} \\
    \hline % 替换为 \hline
    \end{tabular}
    \caption{Performance comparisons using \textbf{DiT-XL} models on ImageNet dataset using 50 denoising steps. }
    \label{tab:dit_full_final}
\end{table}
\paragraph{\textbf{Distillation Training Performance}}
\begin{figure}[htbp]
    \centering
    \includegraphics[width=1.0\textwidth]{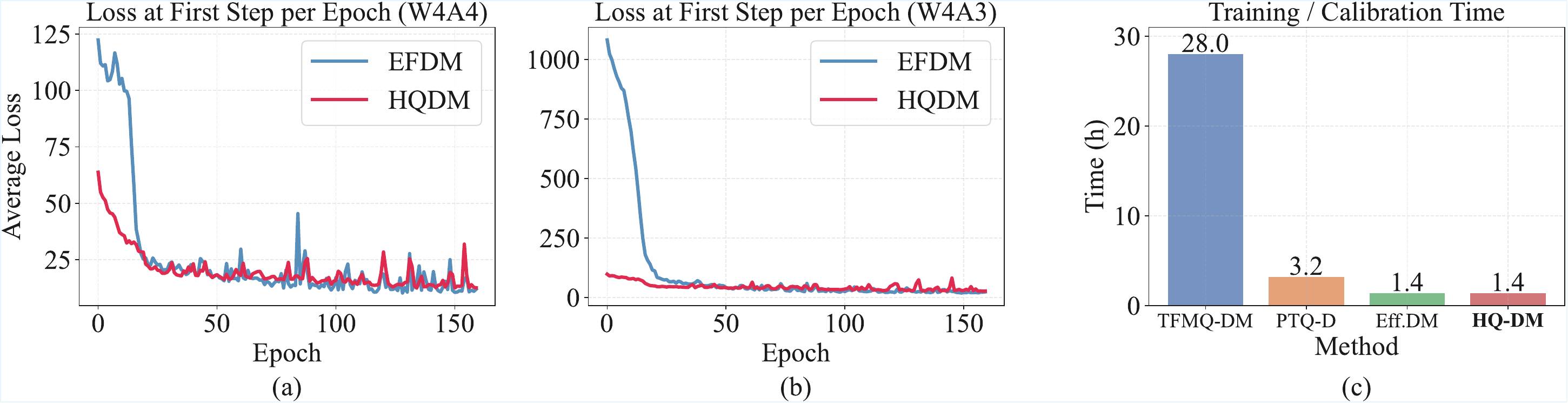}
    \caption{(a-b) Training loss curves for W4A4 and W4A3 configurations on the ImageNet $256\times256$ dataset. We observe the loss change at step 0, as the first step of the training process corresponds to the final step of the inference denoising output. (c) Comparison of time costs for training or calibration across different quantization methods.}
    \label{fig:loss_curve}
\end{figure}
Figure~\ref{fig:loss_curve} (a) (b) illustrates the convergence characteristics of HQ-DM under W4A4 and W4A3 distillation training settings. The results indicate that HQ-DM achieves more stable convergence compared to EfficientDM, with accelerated convergence rates. As the activation quantization bit-width decreases, the acceleration and stability of training become increasingly pronounced. Figure~\ref{fig:loss_curve} (c) compares the calibration and training time costs of various quantization schemes on a single A100 GPU. The results demonstrate that HQ-DM achieves a training efficiency comparable to EfficientDM, with both requiring significantly lower time costs. This further indicates that the computational overhead introduced by the sparse Hadamard transformation (i.e. $\mathbf{H}_k^{\text{raw}}$) is negligible. Consequently, this underscores that HQ-DM is a highly efficient quantization solution.

\subsection{Ablation Study}
\paragraph{\textbf{Component Study}}
\begin{table}[!htb]
    \centering
    \setlength{\tabcolsep}{3pt}
    \begin{tabular}{lcccccccc}
    \hline
        Method/Variant & LSQ & LoRA & Hada. & IS$\uparrow$ & FID$\downarrow$ & sFID$\downarrow$ & Prec.(\%)$\uparrow$ \\
    \hline
        FP32 Baseline &  &  &  & 365.35 & 11.22 & 7.78 & 93.68 \\
        EfficientDM & \checkmark & \checkmark &  & 220.20 & 6.63 & 9.80 & 80.97 \\
    \hline
        Naive PTQ + LSQ & \checkmark &  &  & 5.09 & 193.59 & 67.24 & 3.38 \\
        Naive PTQ + LSQ + Hada. & \checkmark &  & \checkmark & \textbf{33.29} & \textbf{68.16} & \textbf{48.86} & \textbf{26.50} \\
        Naive QAT + LSQ & \checkmark &  &  & 159.24 & 7.69 & 8.38 & 76.85 \\
        Naive QAT + LSQ + Hada. & \checkmark &  & \checkmark & \textbf{203.69} & \textbf{6.47} & 10.98 & \textbf{82.53} \\
    \hline
        LoRA only &  & \checkmark &  & 185.51 & 8.55 & 10.46 & 73.25 \\
        LoRA + Hada. &  & \checkmark & \checkmark & \textbf{229.75} & \textbf{6.52} & 10.72 & \textbf{79.55} \\
    \hline
        \textbf{HQ-DM (Full)} & \checkmark & \checkmark & \checkmark & 248.30 & 6.58 & 9.60 & 84.44 \\
    \hline
    \end{tabular}
    \caption{W4A4 Ablation study of fully-quantized LDM-4 models on ImageNet $256\times256$. We evaluate the impact of LSQ (timestep-aware learning quantization scales), LoRA (LoRA-based distillation), and our Hadamard Transformation (Hada.).}
    \label{tab:ablation_final}
\end{table}
Table~\ref{tab:ablation_final} presents a component-wise ablation study of HQ-DM on the LDM-4 model (ImageNet $256\times256$) under W4A4 quantization. The evaluation demonstrates that the Hadamard transformation operates independently of distillation adjustments. For example, it significantly enhances the Naive PTQ scheme (where LSQ here denotes the application of timestep-aware static scales) by elevating the IS from 5.09 to 33.29 and lowering the FID by 125.43. Furthermore, the experiments without learnable LSQ confirm that HQ-DM’s gains are not reliant on LSQ, as it maintains robust performance (e.g., 6.52 FID in the LoRA+Hada. variant) even with a fixed scale.
\begin{table}[!t]
    \centering
    \begin{tabular}{lccccccc}
    \hline
        Method & Bit (W/A) & IS $\uparrow$ & FID $\downarrow$ & sFID $\downarrow$ & Precision(\%) $\uparrow$  & \makecell{Latency \\ (s/sample)} $\downarrow$ & Speedup $\uparrow$ \\
    \hline
        FP & 32/32 & 365.35 & 11.22 & 7.78 & 93.68 & - & - \\
    \hline
        DHQ-DM & 8/8 & 5.72 & 178.07 & 36.17 & 24.92 & 1.0688 & 1.00$\times$ \\
        \textbf{HQ-DM} & 8/8 & \textbf{363.22} & \textbf{11.01} & \textbf{7.76} & \textbf{93.46} & \textbf{0.9419} & \textbf{1.13$\times$} \\
    \hline
        DHQ-DM & 4/8 & 6.13 & 199.00 & 35.48 & 14.41 & 1.1510 & 1.00$\times$ \\
        \textbf{HQ-DM} & 4/8 & \textbf{355.59} & \textbf{10.07} & \textbf{7.14} & \textbf{93.07} & \textbf{1.0257} & \textbf{1.12$\times$} \\
    \hline
        DHQ-DM & 8/4 & 4.74 & 167.97 & 42.42 & 18.75 & 1.0677 & 1.00$\times$ \\
        \textbf{HQ-DM} & 8/4 & \textbf{292.76} & \textbf{7.68} & \textbf{9.93} & \textbf{88.68} & \textbf{0.9391} & \textbf{1.14$\times$} \\
    \hline
        DHQ-DM & 4/4 & 4.37 & 179.40 & 49.20 & 14.63 & 1.1585 & 1.00$\times$ \\
        \textbf{HQ-DM} & 4/4 & \textbf{248.30} & \textbf{6.58} & \textbf{9.60} & \textbf{84.44} & \textbf{1.0581} & \textbf{1.10$\times$} \\
    \hline
    \end{tabular}
    \caption{Performance and efficiency comparisons between DHQ-DM (Double Hadamard Transformation) and HQ-DM on ImageNet $256\times256$. Latency is measured per sample. Speedup is calculated relative to DHQ-DM within each bit-width.}
    \label{tab:double_single}
\end{table}
\paragraph{\textbf{Transformation Scheme}}
Table~\ref{tab:double_single} provides a detailed comparison between the traditional Double Hadamard Transformation and the Single Hadamard Transformation using the LDM-4 model on ImageNet $256\times256$ dataset. Since Double Hadamard Transformation is inherently incompatible with convolutions, these layers are regressed to the Single Hadamard Transformation for a fair comparison. It is observed that the Double Hadamard Transformation consistently lags in performance across all quantization settings, with an inference latency overhead exceeding 10\%. This performance degradation aligns with our earlier assertion: imposing Hadamard transformation on weights containing minimal outliers can exacerbate quantization noise, which then propagates through the iterative sampling process. Additional ablation results are provided in the Appendix.

\section{Conclusion}
We propose HQ-DM, a novel framework that enables low-bit quantization for diffusion models. To address the outlier issue in activation values during diffusion model computation, HQ-DM employs a Single Hadamard Transformation to reduce activation outliers, thereby enhancing quantized model performance. Furthermore, this hardware-friendly framework is capable of supporting integer convolution operations without amplifying outliers in the weights during computation, thereby enhancing performance.

\clearpage
\bibliographystyle{splncs04}
\bibliography{main}

@String(CVPR  = {IEEE Conf. Comput. Vis. Pattern Recog.})

@String(ICCV  = {Int. Conf. Comput. Vis.})

@String(NeurIPS = {Adv. Neural Inform. Process. Syst.})

@String(ICML  = {Int. Conf. Mach. Learn.})

@String(ICLR  = {Int. Conf. Learn. Represent.})

@String(AAAI  = {AAAI})

@String(CVPR  = {CVPR})

@String(ICCV  = {ICCV})

@String(NeurIPS = {NeurIPS})

@String(ICML  = {ICML})

@String(ICLR  = {ICLR})

@inproceedings{Ho2020DDPM,
  author    = {Jonathan Ho and Ajay Jain and Pieter Abbeel},
  title     = {Denoising Diffusion Probabilistic Models},
  booktitle = {Advances in Neural Information Processing Systems (NeurIPS)},
  volume = {33},
  pages     = {6840--6851},
  year      = {2020}
}

@inproceedings{Karras2019StyleGAN,
  author    = {Tero Karras and Samuli Laine and Timo Aila},
  title     = {A Style-Based Generator Architecture for Generative Adversarial Networks},
  booktitle = {IEEE/CVF Conference on Computer Vision and Pattern Recognition (CVPR)},
  pages     = {4401--4410},
  year      = {2019}
}

@inproceedings{loshchilovdecoupled,
  title={Decoupled Weight Decay Regularization},
  author={Loshchilov, Ilya and Hutter, Frank},
  booktitle={International Conference on Learning Representations},
  year= {2019},
}

@inproceedings{Rombach2022LDM,
  author    = {Robin Rombach and Andreas Blattmann and Dominik Lorenz and Patrick Esser and Bj{\"o}rn Ommer},
  title     = {High-Resolution Image Synthesis with Latent Diffusion Models},
  booktitle = {IEEE/CVF Conference on Computer Vision and Pattern Recognition (CVPR)},
  year      = {2022}
}

@article{yu2015lsun,
  title={{LSUN:} Construction of a large-scale image dataset using deep learning with humans in the loop},
  author={Yu, Fisher and Seff, Ari and Zhang, Yinda and Song, Shuran and Funkhouser, Thomas and Xiao, Jianxiong},
  journal={arXiv preprint arXiv:1506.03365},
  year={2015}
}

@article{heusel2017gans,
  title={{GANs} Trained by a Two Time-Scale Update Rule Converge to a Local Nash Equilibrium},
  author={Heusel, Martin and Ramsauer, Hubert and Unterthiner, Thomas and Nessler, Bernhard and Hochreiter, Sepp},
  journal={Advances in neural information processing systems},
  volume={30},
  pages={6626--6637},
  year={2017}
}

@inproceedings{kingma2015adam,
  title={Adam: A method for stochastic optimization},
  author={Diederik P. Kingma and Jimmy Ba},
  booktitle={International Conference on Learning Representations},
  year={2015},
}

@inproceedings{nash2021generating,
  title={Generating images with sparse representations},
  author={Nash, Charlie and Menick, Jacob and Dieleman, Sander and Battaglia, Peter},
  booktitle={International Conference on Machine Learning},
  pages={7958--7968},
  year={2021},
  organization={PMLR}
}

@inproceedings{deng2009imagenet,
  title={Imagenet: A large-scale hierarchical image database},
  author={Deng, Jia and Dong, Wei and Socher, Richard and Li, Li-Jia and Li, Kai and Fei-Fei, Li},
  booktitle={2009 IEEE conference on computer vision and pattern recognition},
  pages={248--255},
  year={2009}
}

@article{salimans2016improved,
  title={Improved techniques for training gans},
  author={Salimans, Tim and Goodfellow, Ian and Zaremba, Wojciech and Cheung, Vicki and Radford, Alec and Chen, Xi},
  journal={Advances in neural information processing systems},
  volume={29},
  pages={2226--2234},
  year={2016}
}

@inproceedings{weiqdrop,
  title={QDrop: Randomly Dropping Quantization for Extremely Low-bit Post-Training Quantization},
  author={Wei, Xiuying and Gong, Ruihao and Li, Yuhang and Liu, Xianglong and Yu, Fengwei},
  year={2022},
  booktitle={International Conference on Learning Representations}
}

@article{dettmers2023qlora,
  title={Qlora: Efficient finetuning of quantized llms},
  author={Dettmers, Tim and Pagnoni, Artidoro and Holtzman, Ari and Zettlemoyer, Luke},
  journal={Advances in neural information processing systems},
  volume={36},
  pages={10088--10115},
  year={2023}
}

@inproceedings{
    hu2022lora,
    title={Lo{RA}: Low-Rank Adaptation of Large Language Models},
    author={Edward J Hu and Yelong Shen and Phillip Wallis and Zeyuan Allen-Zhu and Yuanzhi Li and Shean Wang and Lu Wang and Weizhu Chen},
    booktitle={International Conference on Learning Representations},
    year={2022}
}

@inproceedings{podell2024sdxl,
  title={{SDXL:} Improving latent diffusion models for high-resolution image synthesis},
  author={Podell, Dustin and English, Zion and Lacey, Kyle and Blattmann, Andreas and Dockhorn, Tim and M{\"u}ller, Jonas and Penna, Joe and Rombach, Robin},
  booktitle={International Conference on Learning Representations},
  year={2024}
}

@inproceedings{Oord2016WaveNet,
  author    = {Aaron van den Oord and Sander Dieleman and Heiga Zen and Karen Simonyan and Oriol Vinyals and Alex Graves and Nal Kalchbrenner and Andrew Senior and Koray Kavukcuoglu},
  title     = {WaveNet: A Generative Model for Raw Audio},
  booktitle = {Speech Synthesis Workshop (SSW)},
  year      = {2016}
}

@inproceedings{shen2024naturalspeech,
  title={NaturalSpeech 2: Latent Diffusion Models are Natural and Zero-Shot Speech and Singing Synthesizers},
  author={Kai Shen and Zeqian Ju and Xu Tan and Eric Liu and Yichong Leng and Lei He and Tao Qin and Sheng Zhao and Jiang Bian},
  booktitle={International conference on learning representations},
  year={2024}
}

@article{liu2024audioldm,
  title={Audioldm 2: Learning holistic audio generation with self-supervised pretraining},
  author={Liu, Haohe and Yuan, Yi and Liu, Xubo and Mei, Xinhao and Kong, Qiuqiang and Tian, Qiao and Wang, Yuping and Wang, Wenwu and Wang, Yuxuan and Plumbley, Mark D},
  journal={IEEE/ACM Transactions on Audio, Speech, and Language Processing},
  volume={32},
  pages={2871--2883},
  year={2024},
  publisher={IEEE}
}

@inproceedings{Brown2020GPT3,
  author    = {Tom B. Brown and Benjamin Mann and Nick Ryder and Melanie Subbiah and Jared Kaplan and Prafulla Dhariwal and Arvind Neelakantan and Pranav Shyam and Girish Sastry and Amanda Askell and Sandhini Agarwal and Ariel Herbert{-}Voss and Gretchen Krueger and Tom Henighan and Rewon Child and Aditya Ramesh and Daniel M. Ziegler and Jeffrey Wu and Clemens Winter and Christopher Hesse and Mark Chen and Eric Sigler and Mateusz Litwin and Scott Gray and Benjamin Chess and Jack Clark and Christopher Berner and Sam McCandlish and Alec Radford and Ilya Sutskever and Dario Amodei},
  title     = {Language Models are Few-Shot Learners},
  booktitle = {Advances in Neural Information Processing Systems (NeurIPS)},
  year      = {2020}
}

@article{touvron2023llama,
  title={Llama: Open and efficient foundation language models},
  author={Touvron, Hugo and Lavril, Thibaut and Izacard, Gautier and Martinet, Xavier and Lachaux, Marie-Anne and Lacroix, Timoth{\'e}e and Rozi{\`e}re, Baptiste and Goyal, Naman and Hambro, Eric and Azhar, Faisal and others},
  journal={arXiv preprint arXiv:2302.13971},
  year={2023}
}

@inproceedings{Song2021DDIM,
  author    = {Jiaming Song and Chenlin Meng and Stefano Ermon},
  title     = {Denoising Diffusion Implicit Models},
  booktitle = {International Conference on Learning Representations (ICLR)},
  year      = {2021}
}

@inproceedings{Li2023QDiffusion,
  author    = {Xiuyu Li and Yijiang Liu and Long Lian and Huanrui Yang and Zhen Dong and Daniel Kang and Shanghang Zhang and Kurt Keutzer},
  title     = {Q-Diffusion: Quantizing Diffusion Models},
  booktitle = {Proceedings of the IEEE/CVF International Conference on Computer Vision (ICCV)},
  year      = {2023}
}

@inproceedings{Ryu2025DGQ,
  author    = {Hyogon Ryu and NaHyeon Park and Hyunjung Shim},
  title     = {{DGQ:} Distribution-Aware Group Quantization for Text-to-Image Diffusion Models},
  booktitle = {International Conference on Learning Representations (ICLR)},
  year      = {2025}
}

@inproceedings{Gao2025MoDiff,
  author    = {Weizhi Gao and Zhichao Hou and Junqi Yin and Feiyi Wang and Linyu Peng and Xiaorui Liu},
  title     = {Modulated Diffusion: Accelerating Generative Modeling with Modulated Quantization},
  booktitle = {Proceedings of the 42nd International Conference on Machine Learning (ICML)},
  year      = {2025}
}

@inproceedings{librecq,
  title={{BRECQ:} Pushing the Limit of Post-Training Quantization by Block Reconstruction},
  author={Li, Yuhang and Gong, Ruihao and Tan, Xu and Yang, Yang and Hu, Peng and Zhang, Qi and Yu, Fengwei and Wang, Wei and Gu, Shi},
  year= {2021},
  booktitle={International Conference on Learning Representations}
}

@inproceedings{Esser2020LSQ,
  author    = {Steven K. Esser and Jeffrey L. McKinstry and Deepika Bablani and Rathinakumar Appuswamy and Dharmendra S. Modha},
  title     = {Learned Step Size Quantization},
  booktitle = {International Conference on Learning Representations (ICLR)},
  year      = {2020}
}

@article{zhu2024survey,
  title={A survey on model compression for large language models},
  author={Zhu, Xunyu and Li, Jian and Liu, Yong and Ma, Can and Wang, Weiping},
  journal={Transactions of the Association for Computational Linguistics},
  volume={12},
  pages={1556--1577},
  year={2024},
  publisher={MIT Press 255 Main Street, 9th Floor, Cambridge, Massachusetts 02142, USA}
}

@inproceedings{nichol2021improved,
  title={Improved denoising diffusion probabilistic models},
  author={Nichol, Alexander Quinn and Dhariwal, Prafulla},
  booktitle={International conference on machine learning},
  pages={8162--8171},
  year={2021},
  organization={PMLR}
}

@inproceedings{So2023TDQ,
  author = {Junhyuk So and Jungwon Lee and Daehyun Ahn and Hyungjun Kim and Eunhyeok Park},
  title     = {Temporal Dynamic Quantization for Diffusion Models},
  booktitle = {Advances in Neural Information Processing Systems (NeurIPS)},
  year      = {2023}
}

@inproceedings{shang2023post,
  title={Post-training quantization on diffusion models},
  author={Shang, Yuzhang and Yuan, Zhihang and Xie, Bin and Wu, Bingzhe and Yan, Yan},
  booktitle={Proceedings of the IEEE/CVF conference on computer vision and pattern recognition},
  pages={1972--1981},
  year={2023}
}

@inproceedings{he2023ptqd,
  author    = {Yefei He and Luping Liu and Jing Liu and Weijia Wu and Hong Zhou and Bohan Zhuang},
  title     = {{PTQD:} Accurate Post‑Training Quantization for Diffusion Models},
  booktitle = {Advances in Neural Information Processing Systems (NeurIPS)},
  year      = {2023}
}

@inproceedings{huang2024tfmqdm,
  author    = {Yushi Huang and Ruihao Gong and Jing Liu and Tianlong Chen and Xianglong Liu},
  title     = {{TFMQ‑DM:} Temporal Feature Maintenance Quantization for Diffusion Models},
  booktitle = {IEEE/CVF Conference on Computer Vision and Pattern Recognition (CVPR)},
  pages     = {7362--7371},
  year      = {2024}
}

@inproceedings{wang2024apqdm,
  author    = {Changyuan Wang and Ziwei Wang and Xiuwei Xu and Yansong Tang and Jie Zhou and Jiwen Lu},
  title     = {Towards Accurate Post‑training Quantization for Diffusion Models},
  booktitle = {Proceedings of the IEEE/CVF Conference on Computer Vision and Pattern Recognition (CVPR)},
  year      = {2024},
  pages     = {16026--16035}
}

@inproceedings{wang2025quest,
  title={Quest: Low-bit diffusion model quantization via efficient selective finetuning},
  author={Wang, Haoxuan and Shang, Yuzhang and Yuan, Zhihang and Wu, Junyi and Yan, Junchi and Yan, Yan},
  booktitle={Proceedings of the IEEE/CVF International Conference on Computer Vision},
  pages={15542--15551},
  year={2025}
}

@article{dettmers2022gpt3,
  title={GPT3.int8 (): 8-bit matrix multiplication for transformers at scale},
  author={Dettmers, Tim and Lewis, Mike and Belkada, Younes and Zettlemoyer, Luke},
  journal={Advances in neural information processing systems},
  year={2022}
}

@article{ashkboos2024quarot,
  title={Quarot: Outlier-free 4-bit inference in rotated llms},
  author={Ashkboos, Saleh and Mohtashami, Amirkeivan and Croci, Maximilian L and Li, Bo and Cameron, Pashmina and Jaggi, Martin and Alistarh, Dan and Hoefler, Torsten and Hensman, James},
  journal={Advances in Neural Information Processing Systems},
  volume={37},
  pages={100213--100240},
  year={2024}
}

@article{yao2022zeroquant,
  title={{ZeroQuant}: Efficient and affordable post-training quantization for large-scale transformers},
  author={Yao, Zhewei and Yazdani Aminabadi, Reza and Zhang, Minjia and Wu, Xiaoxia and Li, Conglong and He, Yuxiong},
  journal={Advances in neural information processing systems},
  volume={35},
  pages={27168--27183},
  year={2022}
}

@inproceedings{he2024efficientdm,
  author    = {Yefei He and Jing Liu and Weijia Wu and Hong Zhou and Bohan Zhuang},
  title     = {EfficientDM: Efficient Quantization‑Aware Fine‑Tuning of Low‑Bit Diffusion Models},
  booktitle = {International Conference on Learning Representations (ICLR)},
  year      = {2024}
}

@article{wei2022outlier,
  title={Outlier suppression: Pushing the limit of low-bit transformer language models},
  author={Wei, Xiuying and Zhang, Yunchen and Zhang, Xiangguo and Gong, Ruihao and Zhang, Shanghang and Zhang, Qi and Yu, Fengwei and Liu, Xianglong},
  journal={Advances in Neural Information Processing Systems},
  volume={35},
  pages={17402--17414},
  year={2022}
}

@inproceedings{zhao2019improving,
  title={Improving neural network quantization without retraining using outlier channel splitting},
  author={Zhao, Ritchie and Hu, Yuwei and Dotzel, Jordan and De Sa, Chris and Zhang, Zhiru},
  booktitle={International conference on machine learning},
  pages={7543--7552},
  year={2019},
  organization={PMLR}
}

@inproceedings{xiao2023smoothquant,
  title={{SmoothQuant}: Accurate and efficient post-training quantization for large language models},
  author={Xiao, Guangxuan and Lin, Ji and Seznec, Mickael and Wu, Hao and Demouth, Julien and Han, Song},
  booktitle={International conference on machine learning},
  pages={38087--38099},
  year={2023},
  organization={PMLR}
}

@article{xi2023training,
  title={Training transformers with 4-bit integers},
  author={Xi, Haocheng and Li, Changhao and Chen, Jianfei and Zhu, Jun},
  journal={Advances in Neural Information Processing Systems},
  volume={36},
  pages={49146--49168},
  year={2023}
}

@inproceedings{peebles2023scalable,
  title={Scalable diffusion models with transformers},
  author={Peebles, William and Xie, Saining},
  booktitle={Proceedings of the IEEE/CVF international conference on computer vision},
  year={2023}
}

@inproceedings{lee2025dmq,
  title={Dmq: Dissecting outliers of diffusion models for post-training quantization},
  author={Lee, Dongyeun and Hur, Jiwan and Shon, Hyounguk and Lee, Jae Young and Kim, Junmo},
  booktitle={Proceedings of the IEEE/CVF International Conference on Computer Vision},
  pages={18510--18520},
  year={2025}
}

@inproceedings{feng2025mpq,
  title={Mpq-dm: Mixed precision quantization for extremely low bit diffusion models},
  author={Feng, Weilun and Qin, Haotong and Yang, Chuanguang and An, Zhulin and Huang, Libo and Diao, Boyu and Wang, Fei and Tao, Renshuai and Xu, Yongjun and Magno, Michele},
  booktitle={Proceedings of the AAAI Conference on Artificial Intelligence},
  volume={39},
  number={16},
  pages={16595--16603},
  year={2025}
}
\end{document}